# On Arrhythmia Detection by Deep Learning and Multidimensional Representation

**K.S. Rajput** BIOFOURMIS
**S. Wibowo** BIOFOURMIS
**C. Hao** BIOFOURMIS
**M. Majmudar** BIOFOURMIS

## Abstract

An electrocardiogram (ECG) is a time-series signal that is represented by one-dimensional (1-D) data. Higher dimensional representation contains more information that is accessible for feature extraction. Hidden variables such as frequency relation and morphology of segment is not directly accessible in the time domain. In this paper, 1-D time series data is converted into multi-dimensional representation in the form of multichannel 2-D images. Following that, deep learning was used to train a deep neural network based classifier to detect arrhythmias. The results of simulation on testing database demonstrate the effectiveness of the proposed methodology by showing an outstanding classification performance compared to other existing methods and hand-crafted annotations made by certified cardiologists.

## 1. Introduction

Computer interpreted ECGs in clinical use today have limitations in diagnostic accuracy (1) and require human review and secondary interpretation. The availability of highly accurate and fully automated ECG analysis using machine learning has the potential to significantly improve the accuracy as well as time and cost associated with clinical ECG interpretation. Given the availability of on-demand ECG diagnostics (e.g., AliveCor. Apple Watch), there is a growing need for fast and accurate automated ECG interpretation techniques.

Recently, the deep learning technique (4) has emerged as a promising solution for ECG interpretation (49). Structured as a multi-layer network (5), deep learning extract features automatically from the original images / signals and optimize the model (weights and gradients) by backpropagation (39). Due to its high performance in automated classification, deep learning based ECG interpretation becomes progressively promising.

Specifically, deep learning has been used to classify heartbeat in (41), where a convolutional neural network is applied, consisting of 3 convolution layers, 3 max-pooling layers, and 3 fully connected layers. On the other hand, convolutional neural network has been used to classify ECG segments (40). The model performs with acceptable performance when classifying between atrial fibrillation, atrial flutter, ventricular fibrillation, and normal sinus rhythm. Beside convolution neural network, deep belief network has been studied to perform classification of ECG (42). The model achieves high accuracy in detecting ventricular ectopic beat and supraventricular ectopic beat. The technique involves extracting the feature based on morphology and RR-interval. The output of that is fed into a multilayer deep neural network. Several aforementioned studies show a promising capability of deep learning on the field of ECG classification.

This paper proposes an enhanced deep learning technique to detect arrhythmias in ECG. As shown in Figure 1, the proposed system collects ECG and converts it by wavelets and short-time Fourier transform for a better representation of beat-to-beat morphologies. Those represented ECG segments is then imported into a dense connected convolutional neural network for arrhythmia classification. As a result, the proposed system becomes more effective and accurate than other existing ECG arrhythmia detection algorithms, and superior than human reviewed annotations made by certified cardiologists.

## 2. Model

The overall methodology of the proposed system is shown in Figure 1, which consists of (1) pre-processing implemented by filtering and segmentation, (2) one-dimensional ECG signal to spectro-temporal images conversion achieved by Morlet wavelet transform and short-time Fourier transform, (3) feature auto-extraction and classification fulfilled by a dense connected convolutional neural network. The following contents introduce those components section by section.

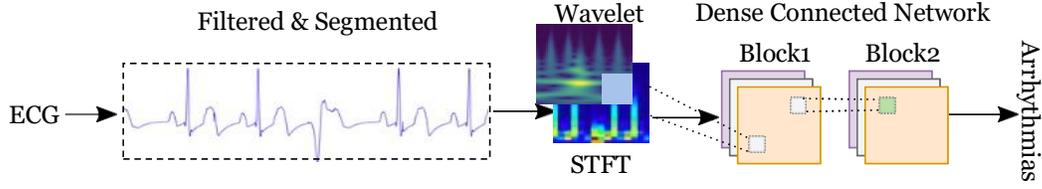

*Figure 1.* The overall methodology of the proposed system for the proposed ECG arrhythmia classification. The system consists of pre-processing (filtering and segmentation), feature extraction (Morlet wavelet transform and short-time Fourier transform) and classification (dense connected network).

**Pre-Processing**

The proposed system takes as input a time-series of raw ECG signal and removes noise through notched filters, bandpass filters, and adaptive filters. The system then split the ECG signal into segments based on a sliding window. Then, those 1D segments are converted to 2D feature images based on wavelets or Fourier transform.

**Feature Extraction**

Wavelet transform (WT) is widely used to describe the ECG morphology features since it's robust to noise and can effectively represent both time and frequency information in different resolutions. In practice, some particular beats which include wide QRS morphologies or more low-frequency spectrums (e.g., bundle branch block beat and ventricular ectopic beat) will be significantly amplified or enlarged by WT while normal beats keep the same. This characteristics of WT distinguishes some special beats and make them easy-to-detect. The proposed system applies Morlet wavelet transform (47) to identify abnormal ECG since it's sensitive to the variance of non-stationary signal and can be considered a proper compensation for Fourier transform.

Short-time Fourier transform (STFT) is another widely used method which describes the real-time signals features in time and frequency domain. STFT uses a Fourier-related transform used to determine the sinusoidal frequency and phase content of local sections of a signal as it changes over time. This proposed system applies STFT to convert the ECG signal into feature images, then perform image classification to identify abnormal ECG. Figure 2 demonstrates the 2D feature images generated from the STFT or Morlet wavelet transform of ECG, which is considered as distinguishable when types of arrhythmia changes.

**Deep Learning Classification**

We use a dense convolutional neural network (48) for the classification task. The high-level architecture of the network is shown in Figure 1. The network consists of residual-like blocks wherein every two, and preceding lay-ers are explicitly connected in a feed-forward fashion.

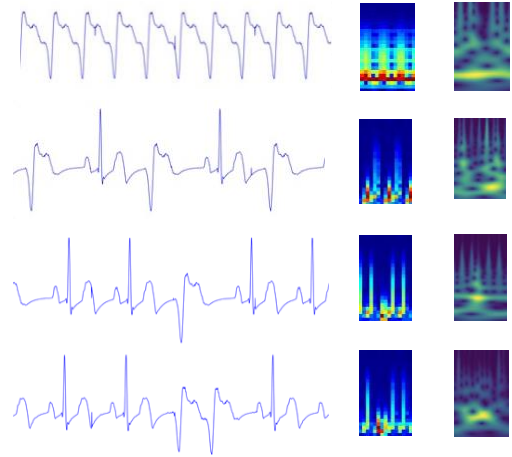

*Figure 2.* original ECG segments and their corresponding spectro-temporal images after processing wavelets transform and short-time Fourier transform.

In each residual-like blocks, direct connections link any layer to all subsequent layers. Assume a single image $x_0$ that is passed through a convolutional network. The network comprises $K$ layers, each of which implements a non-linear transformation $H_k(\ )$, where $k$ indexes the layer. $H_k(\ )$ can be a composite function of operations (the output of $k^{th}$ layer as $x_k$). Consequently, the $k^{th}$ layer receives the feature-maps of all preceding layers, $x_0, \ldots, x_{k-1}$, as input:

$$x_k = H_k([x_0, x_1, \ldots, x_{k-1}]), \qquad (1)$$

where $[x_0, x_1, \ldots, x_{k-1}]$ refers to the concatenation of the feature-maps produced in layers $0, \ldots, k$ -$1$.

Before entering the first residual-like block, a convolution with 16 output channels is performed on the input 2D spectrum feature images. For convolutional layers with kernel size 3x3, each side of the inputs is zero-padded by one pixel to keep the feature-map size fixed. The network uses 1x1 convolution followed by 2x2 average pooling as transition layers between two contiguous residual-like blocks. At the end of the last block, a global average pooling is performed and a softmax classifier is applied. The feature-map sizes in these blocks are 32x32, 16x16, and 8x8, respectively.

## 3. Data Assessment

We used 121,346 single lead episodic ECG records collected from patients with a history of cardiac arrhythmias as the main source of data. Each recording contains 60 second ECG, digitized at 256 samples per second with 11-bit resolution over a 10-mV range. The output of ECG detection is classified into nine groups. Cardiologists annotated the record based on those classes.

Biofourmis' web application was used to annotate the ECG records. The web consists of ECG visualization tool and the dropdown selection to choose the annotation. The system allows to annotate the onset and offset of the rhythm. Besides, each beat was annotated based on the peak location. The annotation system does not allow an overlap of onset and offset rhythm. Moreover, the cardiologists were trained to use the system prior to annotating the ECG records.

We distributed the training data randomly to the cardiologist. We divided the whole data set into 5 batches. For each batch, at least the data was replicated twice. The third cardiologist only annotates the data if there was disagreement between those two cardiologists. We let different batch to be annotated by different annotation so that the work can be finished parallelly.

The annotation that we accept as a golden standard is the one that is agreed by the majority of cardiologists. If there is no disagreement between two cardiologists, then the annotation is accepted. In case there is disagreement between cardiologists, the data will be sent to third cardiologist and the decision is still based on the majority of votes.

There is a possibility where all three cardiologists disagree on a single record. In this case, a different cardiologists group will reannotate again on the same strip to reach the agreement. Besides, the data is rejected and not used if the cardiologist agrees that the data doesn't have good quality or there is wrong lead placement. In this case, the label of each class is not exclusive. An ECG record can be labeled with two different arrhythmias if both of them exist in a strip. Additionally, abnormal beats can be coexistent with the rhythm arrhythmias. In this case, we regard as a true prediction if the model can predict the label based on the majority of voted annotation. Besides, in the case where the majority cannot be reached, true positive is accepted if the model predicts one of the annotations.

For the purpose of training and hyper-parameter optimisation, the original training dataset was split into two parts, i.e., 80% of training set and 20% of validation set. As a result, the proposed deep learning model was trained using the 97,077 single lead episodic ECG records; and it's fine-tuned on 24,269 single lead episodic ECG records.

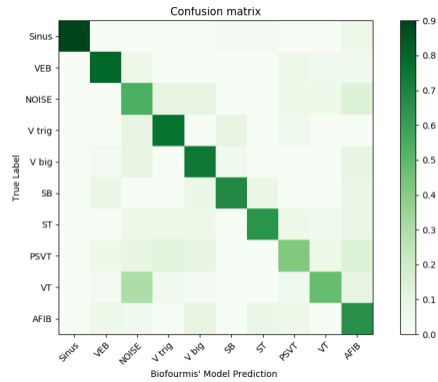

*Figure 3.* A confusion matrix for the model predictions on the testing dataset. Most of arrhythmias can be detected with a proper high accuracy; occasionally noise has a similar waveform or morphology as ventricular tachycardia, leading to a reduction in the result.

For the purpose of testing and performance evaluation, we applied a testing dataset which contains 600 episodic ECG records and covers 30+ arrhythmias. Each record in this dataset contains a 60 second ECG digitized and was annotated by multiple cardiologists in the same way. It should be noted that, the testing dataset is an independent and separated dataset, and it never get involved in any progress of training or hyper-parameter optimisation.

## 4. Discussion

Cardiologists performance was assessed on the testing data. The assessment was done by comparing the majority voted class with the testing annotated data as a golden reference.

There are several previous studies that use deep learning to detect arrhythmias in ECG strip (49). The evaluation method that is used in (49) is by comparing the performance of deep learning model with the average of cardialogists. It has been shown that deep learning with the structure given in (49) can outperform the performance of average cardiologist.

Another study also evaluate deep learning performance to detect arrhythmias in ECG strip (50). The evaluation is done by comparing the model which consist of deep learning structure to Mortara-Veritas in emergency departement. The assessment involves collecting several arrhythmias into different classess or categories. Based on those classess, new metrics of performance can be generated in which the comparison is done with respect to those classes or categories.

|  | F1 scores | | | |
| --- | --- | --- | --- | --- |
|  | Biofourmis' Model | Model(49) | Model(50) | Cardiologists |
| Normal Sinus Rhythm | 0.924 | 0.932 | **0.951** | 0.911 |
| Atrial Fibrillation | **0.838** | 0.697 | 0.752 | 0.724 |
| Sinus Tachycardia | **0.824** | 0.794 | 0.741 | 0.806 |
| Sinus Bradycardia | 0.847 | **0.853** | 0.818 | 0.827 |
| Ventricular Bigeminy | 0.872 | **0.882** | 0.759 | 0.803 |
| Ventricular Trigeminy | **0.880** | 0.855 | 0.731 | 0.780 |
| Ventricular Tachycardia | 0.746 | 0.713 | 0.689 | **0.784** |
| PSVT | **0.716** | 0.618 | 0.602 | 0.654 |
| Noise | **0.779** | 0.707 | 0.632 | 0.713 |
| VEB | **0.909** | 0.872 | 0.824 | 0.834 |
| Summary Results | | | | |
| Specificity | **0.982** | 0.973 | 0.935 | 0.952 |
| Sensitivity | **0.908** | 0.887 | 0.842 | 0.860 |
| F1 | **0.834** | 0.792 | 0.749 | 0.784 |

*Table 1.* Summary of the classification performance

To demonstrate the performance, a confusion matrix is shown in Figure 3 which presents the model prediction results of the proposed system on the testing dataset. As can be shown, most of arrhythmias can be detected with a proper high accuracy. For the exception, the noise might be identified as ventricular tachycardia due to a similar waveform or morphology occasionally.

In addition to demonstrate our performance, we compare our proposed system to the (50) and (49). The paper (50) is the latest ECG deep learning paper which proposed a VGG based convolutional neural network with 16 layers, with the first 13 being convolutional layers, followed by 3 fully connected layers. The paper (49) is a well-known ECG deep learning reference which proposed a residual network based deep learning model with 33 layers of convolution followed by a fully connected layer and a softmax. For comparison, we implement those two deep learning models and compare them to our proposed system on the testing dataset.

Table 1 shows the performance of deep learning model for each different class. The performance was evaluated in terms of F1 scores. In addition, the table also compares the summary of specificity, sensitivity, and F1. The summary was computed by averaging individual performance of each class.

Based on the summary result in table 1, the overall performance of Biofourmis' model outperforms the average of cardiologist and the models that are generated according to previous studies (49) (50). It is expected that the model will have few false alarms due to the presence of arrhythmias that is not covered during the training process. This can be improved by more thorough and complete labeling on different classes of arrhythmias.

## 5. Related Work

An automated ECG classification is an active research topic in the field of machine learning. There has been a lot of application of traditional machine learning to detect arrhythmias in Electrocardiogram (ECG) strip. ECG device records the electrical patterns of the heart's electrical activity which is widely used for automated early diagnosis and better treatment (2). Due to inherent signal quality, commonly the first step is to preprocess the signal such as denoising (31) (32) or baseline removal. It is common practice, before the training phase, the method uses QRS detector (24) to extract ECG's beats (25). After that, feature extraction or dimensional reduction is performed to obtain a meaningful representation of the signal (28) (30). The last step is to train a classifier based on the feature that has been obtained.

One of the issue in ECG classification using machine learning is the performance of the classifier is highly dependent on the accuracy of QRS detector. The error that is produced by QRS detector will propagate to the performance of the classifier. On other hand, the advantage of deep learning comes from its workflow. The workflow of deep learning regards feature extraction, segmentation, and classifier as one paradigm of global training.

Detection engine based on deep learning has consistently been reported to achieve high performance (12) (11). In image classification, deep learning was trained using images from ImageNet (11). It reaches high performance and indicates the viability of deep learning. On the medical field, deep learning is used to classify skin cancer (46). The study (46) report that the performance of deep learning is comparable to dermatologists. Therefore deep learning approach is a promising technique to solve a wide range of problem in medical fields.

# 6. Conclusion

We have trained a deep learning model in which the performance of it is superior for detecting several arrhythmias from ECG records. The main idea that contributes to the performance is the multi-dimensional representation and multilayer deep learning model. It can recover the structure of 1-D data and fit the model based on the training data.

On another aspect, The detection of very rare occurrence arrhythmias remains a very challenging problem. For example, Torsades de pointes has unique shape and feature that deep learning can learn easily, but the small number of available data makes it difficult to train. Additionally, Torsades has a morphology that resembles the shape of noise. It is likely that without enough training data, the deep learning will miss detect Torsades as noise and vice versa.

Automated diagnosis can help clinician and cardiologist to reduce the amount of time spent on analyzing ECG. Furthermore, in the era where connectivity and wearable is affordable, it opens a new possibility where remote analysis can be performed in a place where cardiologist is not accessible.